\pdfoutput=1

\documentclass[11pt]{article}

\usepackage{acl}

\usepackage{times}
\usepackage{latexsym}

\usepackage[T1]{fontenc}

\usepackage[utf8]{inputenc}

\usepackage{microtype}

\usepackage{subfig}
\usepackage{pdfpages}
\usepackage{todonotes}
\usepackage{graphicx}
\usepackage[nolist]{acronym}                
\usepackage{hyperref}                       
\usepackage{xurl}                           
\usepackage{booktabs}                       
\usepackage{multirow}                       
\usepackage{amsfonts}                       
\usepackage{nicefrac}                       
\usepackage{enumitem}                       
\usepackage{pgfplots}                       
\pgfplotsset{compat=1.17} 
\usepackage{pgfplotstable}

\title{ClassActionPrediction: A Challenging Benchmark for Legal Judgment Prediction of Class Action Cases in the US}

\author{
Gil Semo\thanks{\hspace{2mm}Equal Contribution}\hspace{3mm}Dor Bernsohn\hspace{3mm}Ben Hagag\hspace{3mm}Gila Hayat\\
  Darrow AI Ltd.\\
  30 Ha’arbaa Street, Tel Aviv, Israel \\
  \texttt{firstname.lastname@darrow.ai} \\\And
  Joel Niklaus$^*$\thanks{\hspace{2mm}Corresponding Author} \\
  Niklaus.ai\\
  Schwarztorstrasse 108, Bern, Switzerland \\
  \texttt{joel@niklaus.ai} \\
}

\begin{document}
\maketitle

\begin{acronym}[UMLX]
    \acro{FSCS}{Federal Supreme Court of Switzerland}
    \acro{SCI}{Supreme Court of India}
    \acro{ECHR}{European Convention of Human Rights}
    \acro{ECtHR}{European Court of Human Rights}
    \acro{SCOTUS}{Supreme Court of the United States}
    \acro{SPC}{Supreme People's Court of China}
    \acro{SJP}{Swiss-Judgment-Prediction}
    \acro{ASO}{Almost Stochastic Order}
    \acro{ILDC}{Indian Legal Documents Corpus}
    
    \acro{US}{United States}
    \acro{EU}{European Union}

    \acro{NLP}{Natural Language Processing}
    \acro{ML}{Machine Learning}
    \acro{LJP}{Legal Judgment Prediction}
    \acro{SJP}{Swiss-Judgment-Prediction}
    \acro{PJP}{Plea Judgment Prediction}
    
    \acro{BERT}{Bidirectional Encoder Representations from Transformers}

    \acro{CLT}{Cross-Lingual Transfer}
    \acro{HRL}{high resource language}
    \acro{LRL}{low resource language}
    
    \acro{PLM}{Pretrained Language Model}
    \acro{ML}{Machine Learning}
    \acro{NN}{Neural Network}
    
    \acro{TS}{Temperature Scaling}
    \acro{ECE}{Expected Calibration Error}
    \acro{IG}{Integrated Gradients}
    
    \acro{CoA}{Cause of Action}
\end{acronym}

\begin{abstract}
The research field of Legal Natural Language Processing (NLP) has been very active recently, with Legal Judgment Prediction (LJP) becoming one of the most extensively studied tasks.
To date, most publicly released LJP datasets originate from countries with civil law.
In this work, we release, for the first time, a challenging LJP dataset focused on class action cases in the US. It is the first dataset in the common law system that focuses on the harder and more realistic task involving the complaints as input instead of the often used facts summary written by the court.
Additionally, we study the difficulty of the task by collecting expert human predictions, showing that even human experts can only reach 53\% accuracy on this dataset. Our Longformer model clearly outperforms the human baseline (63\%), despite only considering the first 2,048 tokens.
Furthermore, we perform a detailed error analysis and find that the Longformer model is significantly better calibrated than the human experts.
Finally, we publicly release the dataset and the code used for the experiments.
\end{abstract}

\acresetall 

\section{Introduction}

Recently, the literature in Legal \ac{NLP} has grown at a fast pace, firmly establishing it as an important specialized domain in the broader \ac{NLP} ecosystem.
As part of this strong growth and as a first step establishing Legal NLP in the field, many legal datasets have been released in the fields of 
\ac{LJP} \cite{niklaus-etal-2021-swiss, chalkidis_neural_2019}, Law Area Prediction \cite{glaser_classication_2020},
Legal Information Retrieval \cite{wrzalik_gerdalir_2021},
Argument Mining \cite{urchs_design_2022},
Topic Classification \cite{chalkidis_multieurlex_2021},
Named Entity Recognition \cite{luz_de_araujo_lener-br_2018, angelidis_named_2018, leitner_fine-grained_2019},
Natural Language Inference \cite{koreeda_contractnli_2021},
Question Answering \cite{zheng_when_2021, hendrycks_cuad_2021}, and
Summarization \cite{shen_multi-lexsum_2022, kornilova_billsum_2019}.

\begin{figure}[t]
    \centering
    \resizebox{\columnwidth}{!}{
    \includegraphics{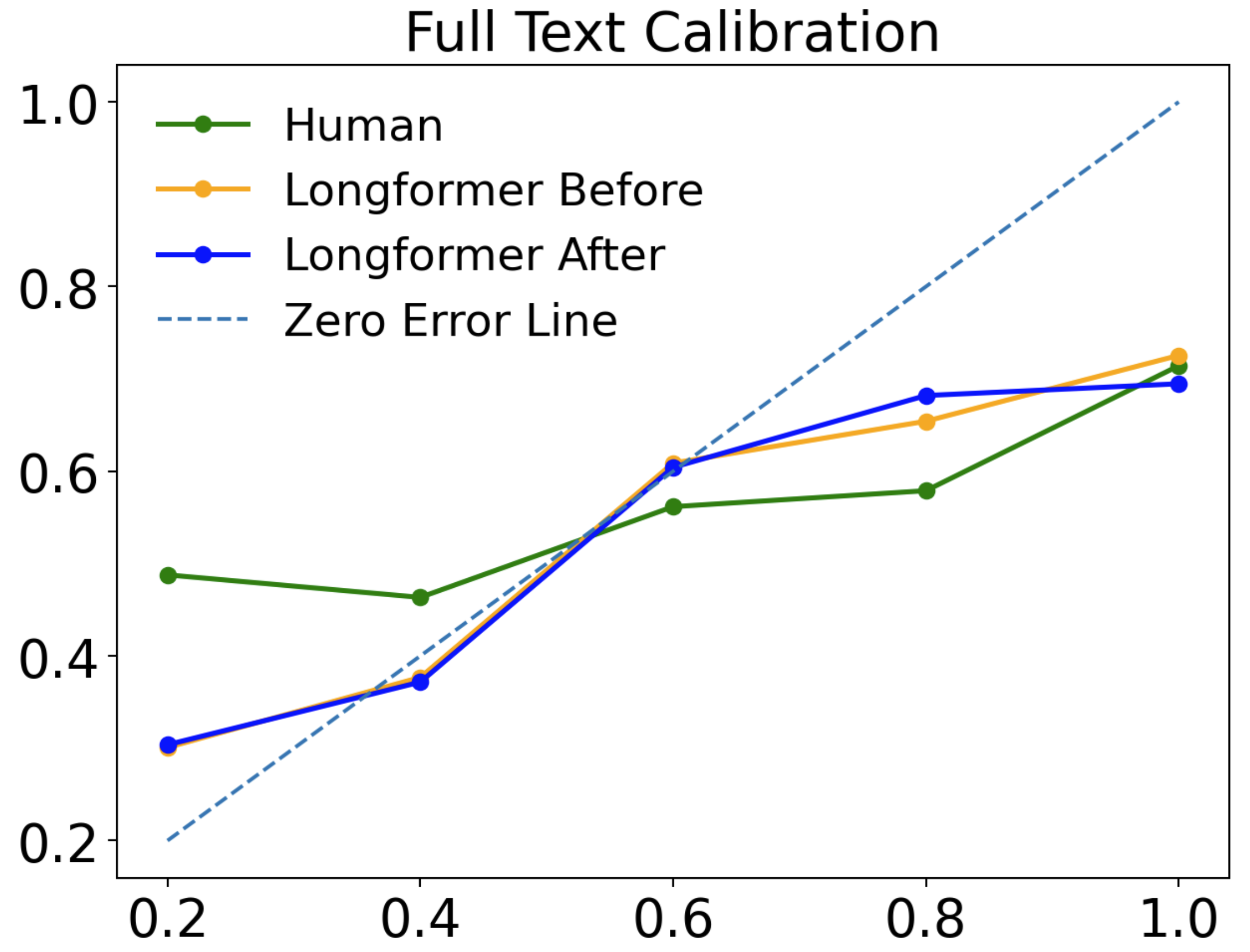}
    }
    \vspace{-8mm}
    \caption{Calibration plot on the Full Text dataset. The human experts rated the confidence of their predictions on a score from 1 to 5. The confidence scores of the Longformer models were binned into 5 buckets.}
    \label{fig:human_calibration}
\vspace{-5mm}
\end{figure}

In particular, the field of \ac{LJP} has been very active, with many datasets released recently. \citet{Cui2022ASO} surveyed the field and divided the datasets into five subtasks. In this work, we release a dataset belonging to the category of the \ac{PJP} task. Most other \ac{PJP} datasets use the facts summary, written by the court (clerks or judges) as input \cite{Cui2022ASO}. The facts are written in such a way as to support the final decision \cite{niklaus-etal-2021-swiss} and require extensive work by highly qualified legal experts \cite{ma_legal_2021}. In contrast, in this work we consider the plaintiff's pleas (AKA complaints) as input, making the task more realistic for use in real-world applications. 

Most \ac{LJP} datasets released so far are from countries with civil law. Our dataset originates from the United States, the largest country employing the common law legal system. To the best of our knowledge, we are the first to release a dataset specifically targeting class action lawsuits.


\subsection*{Motivation} 


The 16th United Nations Sustainable Development Goal (UNSDG) is to “Promote peaceful and inclusive societies for sustainable development, provide access to justice for all and build effective, accountable and inclusive institutions at all levels”. Class actions are a private enforcement instrument that enables courts to organize the mass adjudication of meritorious claims by underrepresented individuals and communities. Without class actions, many victims of illegal action would never get their day in court.
Making case outcomes and facts accessible is crucial to strengthen the effective use of class actions and private enforcement to drive UNSDG 16. With the power of early LJP, plaintiffs will have the ability to bring only meritorious cases to court, and defendants are more likely to resolve them faster.



\subsection*{Main Research Questions}
In this work, we pose and examine three main research questions:

\noindent \textbf{RQ1}: \emph{To what extent is it possible to determine the outcome of US class action cases using only the textual part of the complaints (without metadata)?}

\noindent \textbf{RQ2}: \emph{To what extent can we use \acf{TS} to better calibrate our models?}

\noindent \textbf{RQ3}: \emph{To what extent can expert human lawyers solve the proposed task?}


\subsection*{Contributions}
The contributions of this paper are four-fold:
\vspace{-3mm}
\begin{itemize}[leftmargin=8pt, itemsep=0em]
    \item We curate a new specialized dataset of 10.8K class action complaints in the US from 2012 to 2022 annotated with the binary outcome: win or lose (plaintiff side). In contrast to most other \ac{LJP} datasets it is 
    (a) from a country with the common law system (where there are less datasets available), 
    (b) it is specialized to class actions (important types of complaints ensuring justice for numerous often under-represented individuals), and 
    (c) it uses the plaintiff's pleas as input instead of the facts, making the task more realistic. To the best of our knowledge, our work is the first dataset with plaintiff's pleas in the common law system and in the English language. 
    \item We conduct a detailed analysis of the studied models using \ac{IG} and model calibration using \ac{TS} \cite{pmlr-v70-guo17a}.
    \item We perform an experiment with human experts on a randomly selected subset of the dataset, showing that our Longformer model both outperforms the human experts in terms of accuracy and calibration.
    \item We publicly release a sample of 3,000 cases from the annotated dataset\footnote{\url{https://huggingface.co/datasets/darrow-ai/USClassActions}} together with the human expert labels\footnote{\url{https://huggingface.co/datasets/darrow-ai/USClassActionOutcomes_ExpertsAnnotations}} and the code for the experiments\footnote{\url{https://github.com/darrow-labs/ClassActionPrediction}}.
\end{itemize}


\section{Legal Background}


\subsection{Class Action Lawsuits}
Class actions are a unique procedural instrument that allows one person to sue a company, not only on behalf of himself, but for everyone that has been injured by the same wrongdoing. In contrast to traditional lawsuits, in a class action lawsuit a plaintiff sues the defendant(s) on behalf of a class of absent parties. Class action lawsuits typically involve a minimum of 40 claimants. Rather than filing individual lawsuits for each damaged person, class actions allow the plaintiffs to unite and sue through a single proceeding. Thus, class actions are usually large and important cases and contain more complexity due to the high number of represented plaintiffs. These characteristics make class action a legal enforcement mechanism, along with police and regulators. Class actions both deter companies from harming people in the first place, and give compensation to the large number of victims hurt by the violation, giving consumers power over large corporations.

\subsection{Definitions}


\noindent \textbf{Civil Law vs. Common Law:} In both civil law and common law systems, courts rule based on laws and precedents (previous case law, mostly from the Supreme Court). However, in common law countries (mainly present in the UK and its former Colonies), case law dominates, whereas in civil law countries (most other countries) laws are more important. Note, that the differences are often not clear-cut, and courts usually use a combination of both laws and precedent for their rulings.

\noindent \textbf{Complaint:} 
A complaint is a written pleading to initiate a lawsuit. It includes the plaintiff's cause of action, the court's jurisdiction, and the plaintiff's demand for judicial relief. It is necessary for the complaint to state all of the plaintiff's claims against the defendant, as well as what remedy the plaintiff seeks. A complaint must state ``enough facts to state a claim to relief that is plausible on its face'' \cite{Complaint_stracture}. The standards for filing a complaint vary from state to federal courts, or from one state to another. 
A typical class action complaint contains the allegations, the background details about both the plaintiff and the defendant, and the facts.

\noindent \textbf{Allegations:} 
In a complaint, allegations are statements of claimed facts. These statements are only considered allegations until they are proven. An allegation can be based on information and belief if the person making the statement is unsure of the facts. In the complaint, allegations can appear twice: once as a summary at the beginning and once in more detail later. There is usually a reference to the act that the plaintiff's attorney claims to have been violated in the allegations.

\noindent \textbf{Background Details:}
The complaint contains background sections such as the plaintiff's history, class definitions, the defendant's history, and details about the platform/service in which the allegations took place.

\noindent \textbf{Plaintiff's Facts:} 
The plaintiff's facts or ``factual background'', are statements that can be proven and are often backed up with references and event dates. Note that the plaintiff's facts are written by the plaintiff lawyers.

\noindent \textbf{Facts Summary:} 
The facts summary or ``factual description'', are the summary of the accepted facts by the court and are written by the clerks or judges. The facts summary is usually more condensed in higher courts. Most previous LJP tasks used facts of this type. 
Since in this paper we consider complaints as input, when ``facts'' are mentioned we refer to the plaintiff's facts.

\noindent \textbf{Case Description:} 
The case description is written by the court clerks or judges and usually includes the header, the facts, the considerations, and the rulings.

\subsubsection*{Class Action Outcomes}
Table \ref{tab:binarization} shows the outcomes possible in class action cases. In the following, we briefly describe each of the outcomes.

\noindent \textbf{Settled:} 
``Settling a case'' refers to resolving a dispute before the trial ends.




\noindent \textbf{Uncontested Dismissal:}
Without any opposition from the parties, the case is dismissed and closed. 

\noindent \textbf{Motion to Dismiss:}
The case was dismissed by the court following the defendant’s formal request for a court to dismiss the case. 




\begin{table}[ht]
\centering
\resizebox{\columnwidth}{!}{
\begin{tabular}{lrr}
\toprule
Outcome                         & Bin. Label & \# Examples (\%) \\
\midrule
Settled                         &  win  &    5234 (48.64\%) \\
Other - Plaintiff               &  win  &       58 (00.52\%) \\
Uncontested Dismissal           &  lose &    4544 (42.23\%) \\
Motion to Dismiss               &  lose &     755 (07.01\%) \\
Other - Defendant    &  lose &     170 (01.56\%) \\
\bottomrule
\end{tabular}
}
\caption{This table shows the original outcome together ruled by the court with the frequency and the final binarized label we map it to.}
\label{tab:binarization}
\vspace{-5mm}
\end{table}

\section{Related Work}

\ac{LJP} is an important and well-studied task in legal \ac{NLP}. \citet{Cui2022ASO} subdivide \ac{LJP} into five subtasks: 
(a) In the \emph{Article Recommendation Task}, systems predict relevant law articles for a given case \cite{aletras_predicting_2016,chalkidis_neural_2019,ge_learning_2021}. 
(b) The goal of the \emph{Charge Prediction Task}, mainly studied in China, is to predict the counts the defendant is charged for based on the facts of the case \cite{zhong_legal_2018,hu_few-shot_2018, zhong_iteratively_2020}. 
(c) In the \emph{Prison Term Prediction Task}, systems predict the prison time for the defendant as ruled by the judge \cite{zhong_legal_2018,chen_charge-based_2019}. 
(d) In the \emph{Court View Generation Task}, systems generate court views (explanation written by judges to interpret the judgment decision) \cite{ye_interpretable_2018,wu_-biased_2020}. 
(e) In the \emph{Plea Judgment Prediction Task}, systems predict the case outcome based on the case's facts \cite{niklaus_swiss-judgment-prediction_2021,sulea_predicting_2017,lage-freitas_predicting_2022,long_automatic_2019,ma_legal_2021,strickson_legal_2020,malik_ildc_2021,alali_justice_2021}.
Since our work belongs to the \ac{PJP} category, in the following, we elaborate more on the related work in this area.

\paragraph{Civil Law}
\citet{niklaus_swiss-judgment-prediction_2021} released a trilingual (German, French, Italian) Swiss dataset from the Federal Supreme Court of Switzerland. They use the facts summary 
as input and predict a binary output: approval or dismissal of the plaintiff's pleas for approx. 85K decisions.
\citet{sulea_predicting_2017} released a dataset of approx. 127K French Supreme Court cases. As input, they used the entire case description 
and not only the facts summary, presumably making the task considerably easier and a possible explanation for their high performance on the dataset. As output, they consider up to 8 classes of decisions ruled by the court.
\citet{lage-freitas_predicting_2022} released a dataset comprising roughly 4K cases from a Brazilian State higher court (appellate court). They predicted three labels from the entire case description (written by the judges/clerks).
\citet{jacob_de_menezes-neto_using_2022} release a large dataset of over 765K cases from the 5th Regional Federal Court of Brazil. They investigate a binary prediction task (whether the previous decision was reversed or not) using the entire case description as input.
\citet{long_automatic_2019} studied the \ac{LJP} task on 100K Chinese divorce proceedings considering three types of information as input: applicable law articles, fact description, and plaintiffs' pleas. Their model predicts a binary output.
\citet{ma_legal_2021} released a dataset comprising 70.5K civil cases (private lending) from China. They consider the more realistic task of inputting the plaintiff's complaints (together with debate data) instead of the easier facts summary used by most previous works. As output, their models predict three classes (reject, partially support and support). 
Similarly, our work also studies the more realistic (and challenging) use case of using the plaintiff's pleas as input instead of the heavily processed facts.

\paragraph{Common Law}
\citet{strickson_legal_2020} released a dataset of 5K cases from the UK's highest court of appeal. As input, they consider the case description and their models predict two labels (allow vs. dismiss).
\citet{malik_ildc_2021} study a dataset of 35K Indian Supreme Court cases in English. They use the case description as input and predict a binary outcome (accepted vs. rejected).
\citet{alali_justice_2021} study a dataset of 2.4K US Supreme Court decisions. Their models used the facts summary as input and predicted a binary output (first party won vs. second party won). 
In contrast, our dataset is $\sim$ 5 times larger and is specialized to the rare subset of class action cases.

Apart from \citet{ma_legal_2021}, the \ac{PJP} task based on plaintiff's complaints has not been studied before. In contrast, most previous works studied textual input originating from the case description written by the court.

\section{Dataset Description}

\begin{figure*}[ht]
    \centering
    \resizebox{\textwidth}{!}{
    \subfloat[\centering Top 10 most frequent states]
    {{
    \includegraphics[width=\textwidth/2]{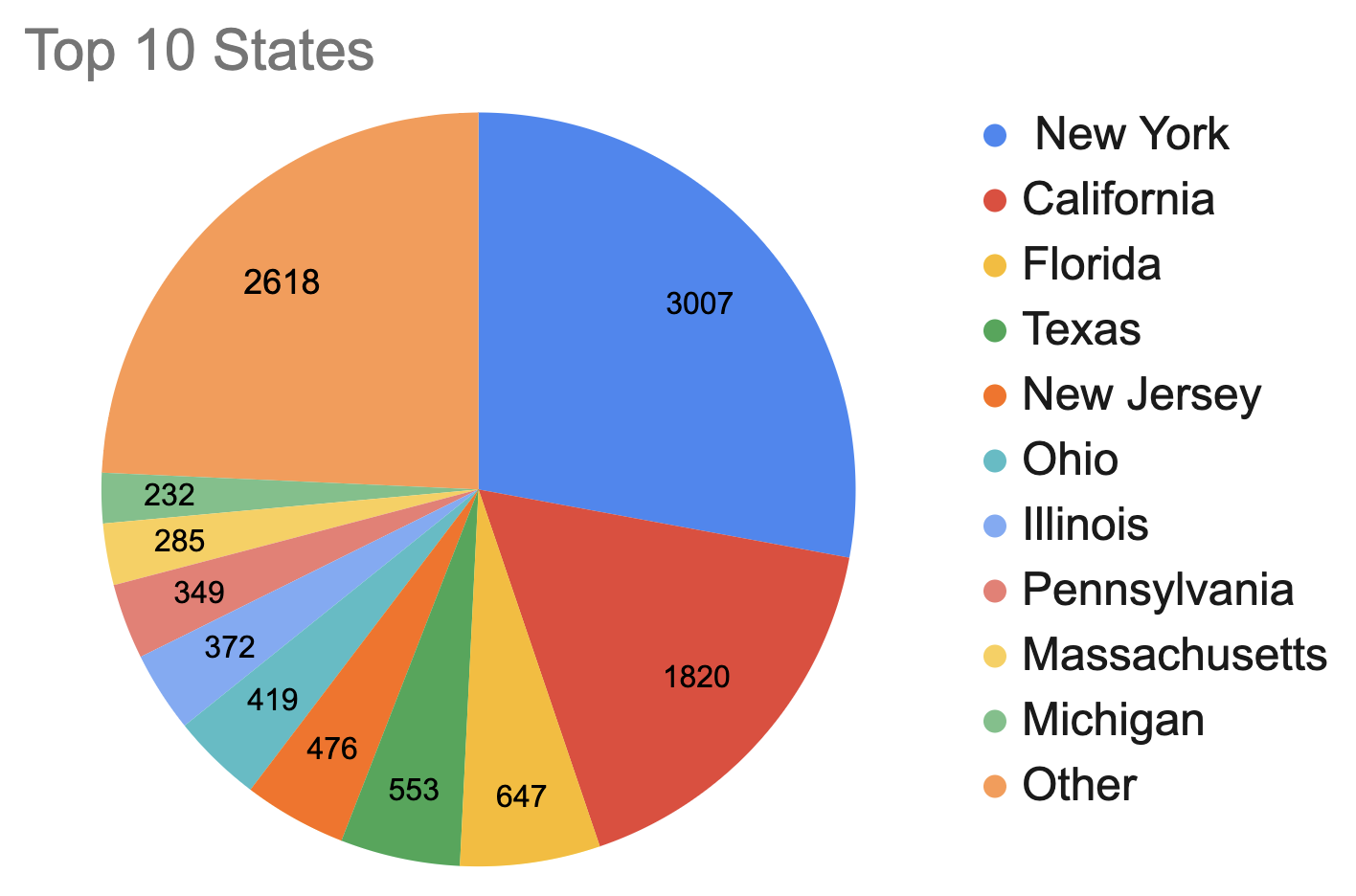} 
    \label{fig:top_10_states}
    }}
    \qquad
    \subfloat[\centering Top 10 most frequent courts]
    {{
    \includegraphics[width=\textwidth/2]{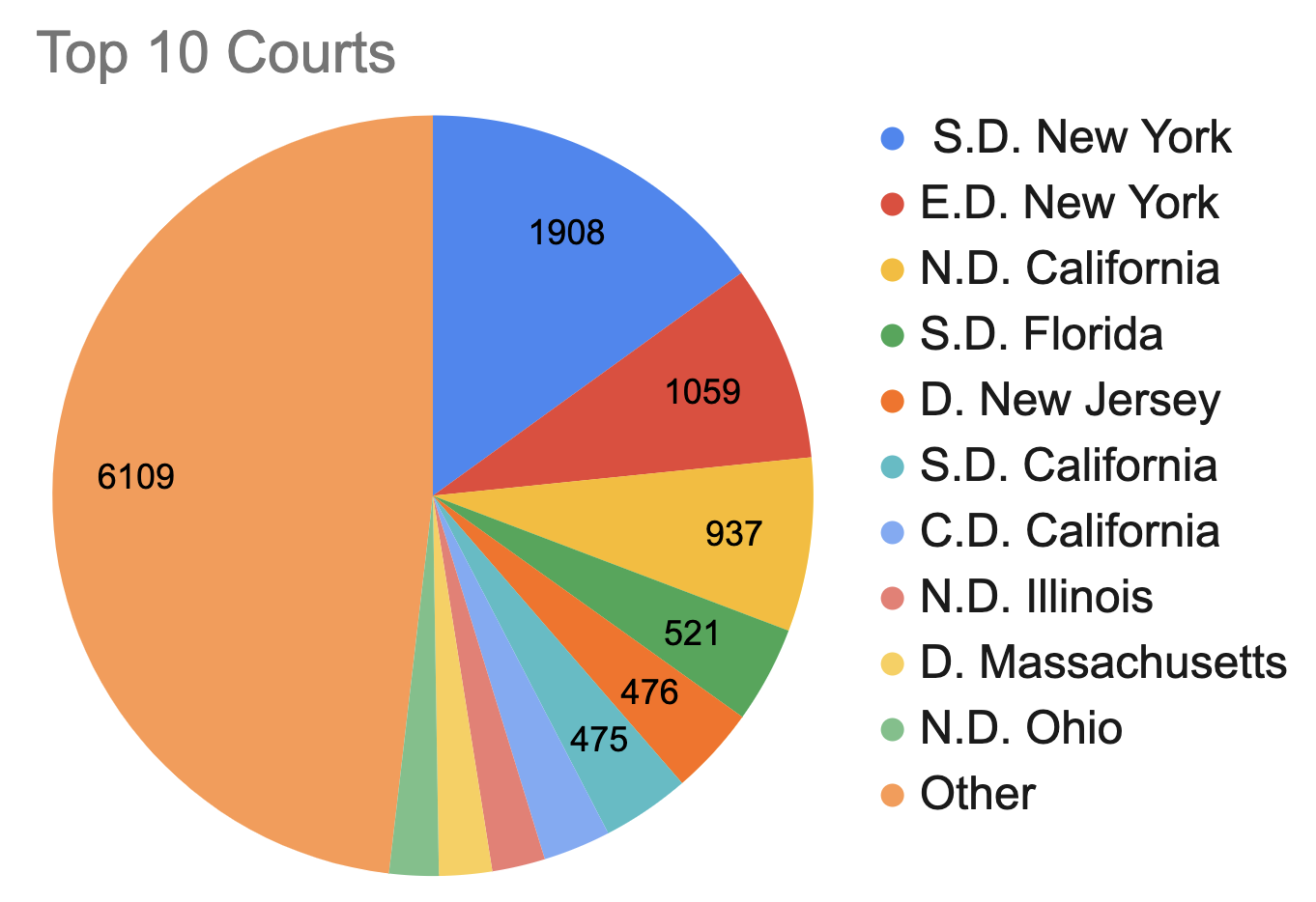} 
    \label{fig:top_10_courts}
    }}
    }
    \caption{Distribution of cases across states and courts. }
    \label{fig:case_distributions}
\end{figure*}


In this section, we describe the dataset origin and statistics in detail. Additionally, we elaborate on the dataset construction process and the variants we produced.

Figures \ref{fig:top_10_states} and \ref{fig:top_10_courts} show the distribution of cases across the most frequent states and courts in the dataset, respectively. Note that the origin of these class action lawsuits is very diverse, both across states and across courts. Not surprisingly, population-rich states like California, Florida, and New York lead the list. However, while California is more than double in population (39.5M vs. 20.2M as of April 2021), the number of class action lawsuits has the inverse relationship ($\sim$ 3K from New York and $\sim$ 1.8K from California). We assume that the complicated filing system in California could be a reason for this disparity\footnote{Each court has its format of filing, and even courts within the same county do not usually use the same complaint filing format.}.

\subsection{Plaintiff's Pleas Instead of Facts Summary}
Condensing and extracting the relevant information from plaintiffs’ pleas and court debates is a large part of the judge's work \cite{ma_legal_2021}. This results in a condensed description of a case's facts. Most previous works consider this condensed description written by the judicial body (judges and clerks) as input. However, since a lot of qualified time has been spent on writing these descriptions, naturally, it makes the \ac{LJP} task easier when using the court-written facts as input. \citet{ma_legal_2021} were the first to consider the original plaintiff's pleas as input on Chinese data. In this work, to the best of our knowledge, we are the first to consider this harder task in the common law system (US class action cases in our case) and in the English language in general.

We do not consider the background details because our models might easily overfit on very specific data. In contrast, our goal was to create a dataset, where models need to focus on case-specific details to solve the task instead of being allowed to consider company-specific information such as number of employees or the area of business. We also disregard the introduction, containing metadata about the judge and the plaintiff.

\subsection{Dataset Construction}

To extract the plaintiffs’ facts and allegations from each case, we manually reviewed hundreds of cases from different courts and different states to learn the structure of the document in each court to build a rule-based regex extraction system that detects the relevant text spans in each complaint. To summarize, constructing the dataset posed many technical difficulties due to the diverse nature of the complaint documents.
At the preprocessing stage, we perform text cleaning, including 
removing some irrelevant text sections that our system incorrectly matched and removing duplicate sections.


\subsection{Label Distribution}


In this work we consider the task of binary legal judgment prediction. To do so, we simplified the labels. We used Table \ref{tab:binarization} to map the outcomes to either \emph{win} or \emph{lose} (for the plaintiff). After binarization the dataset is almost balanced with 5,469 (50.8\%) \emph{lose} cases and 5,290 (49.2\%) \emph{win} cases. Therefore, in our experiments, we just report the accuracy to keep it simple and make the scores more easily interpretable.

\subsection{Dataset Variants}
\label{sec:dataset_variants}

\begin{figure*}[ht]
    \centering
    \resizebox{\textwidth}{!}{
    \subfloat[\centering Full Text]
    {{
    \includegraphics[width=\textwidth/3]{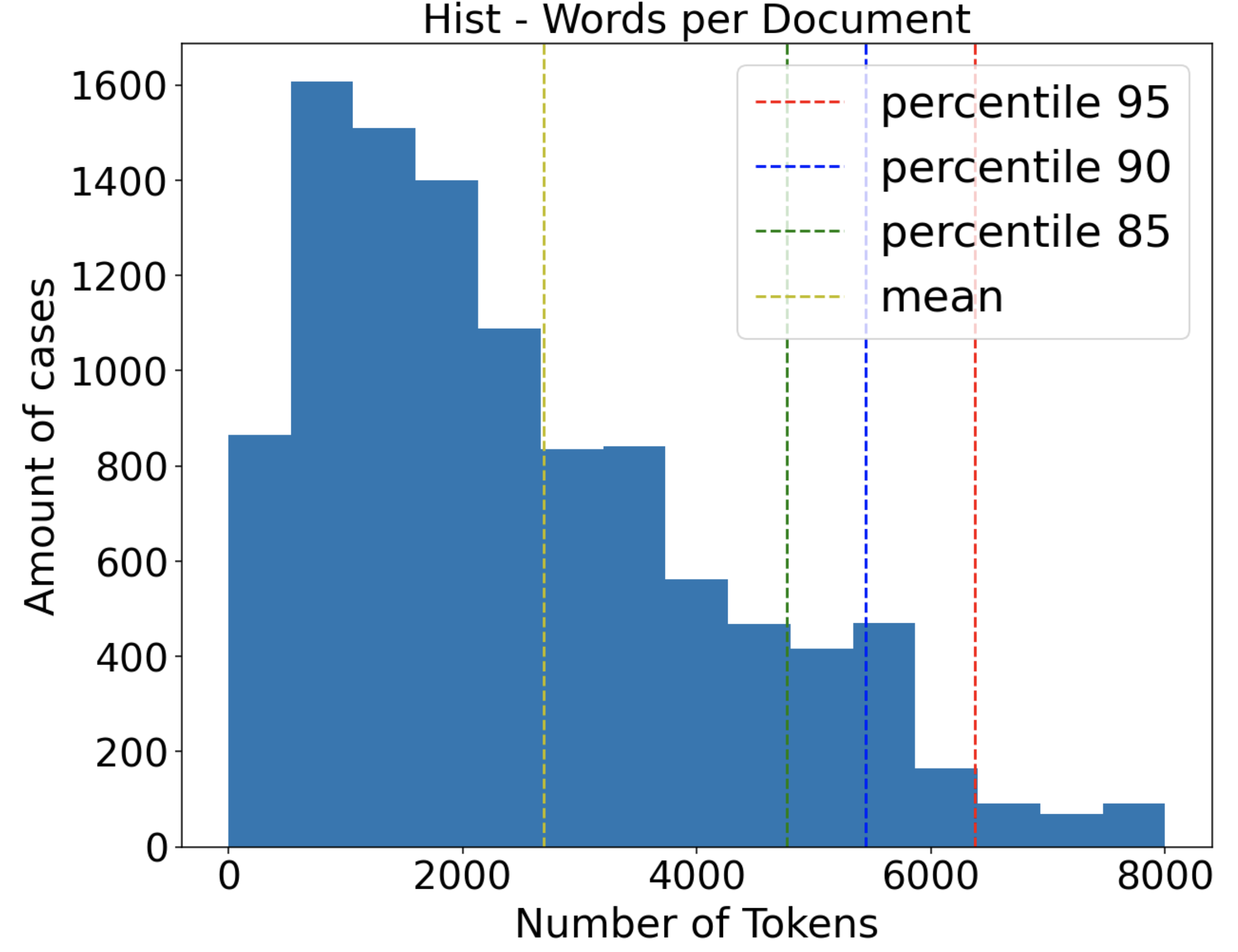} 
    \label{fig:full_text_hist}
    }}
    \qquad
    \subfloat[\centering Unified Allegations]
    {{
    \includegraphics[width=\textwidth/3]{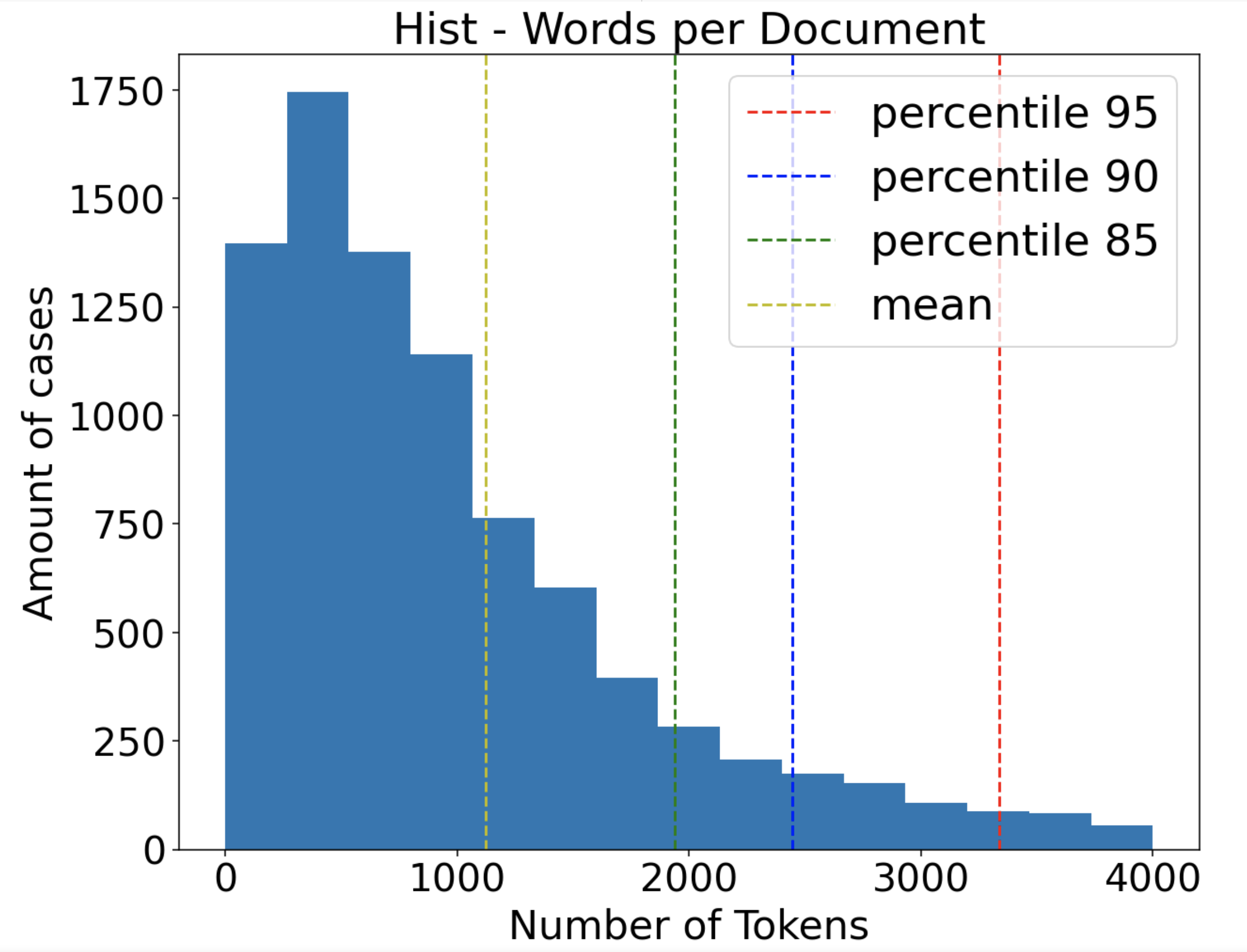} 
    \label{fig:unified_allegations_hist}
    }}
    \qquad
    \subfloat[\centering Separated Allegations]
    {{
    \includegraphics[width=\textwidth/3]{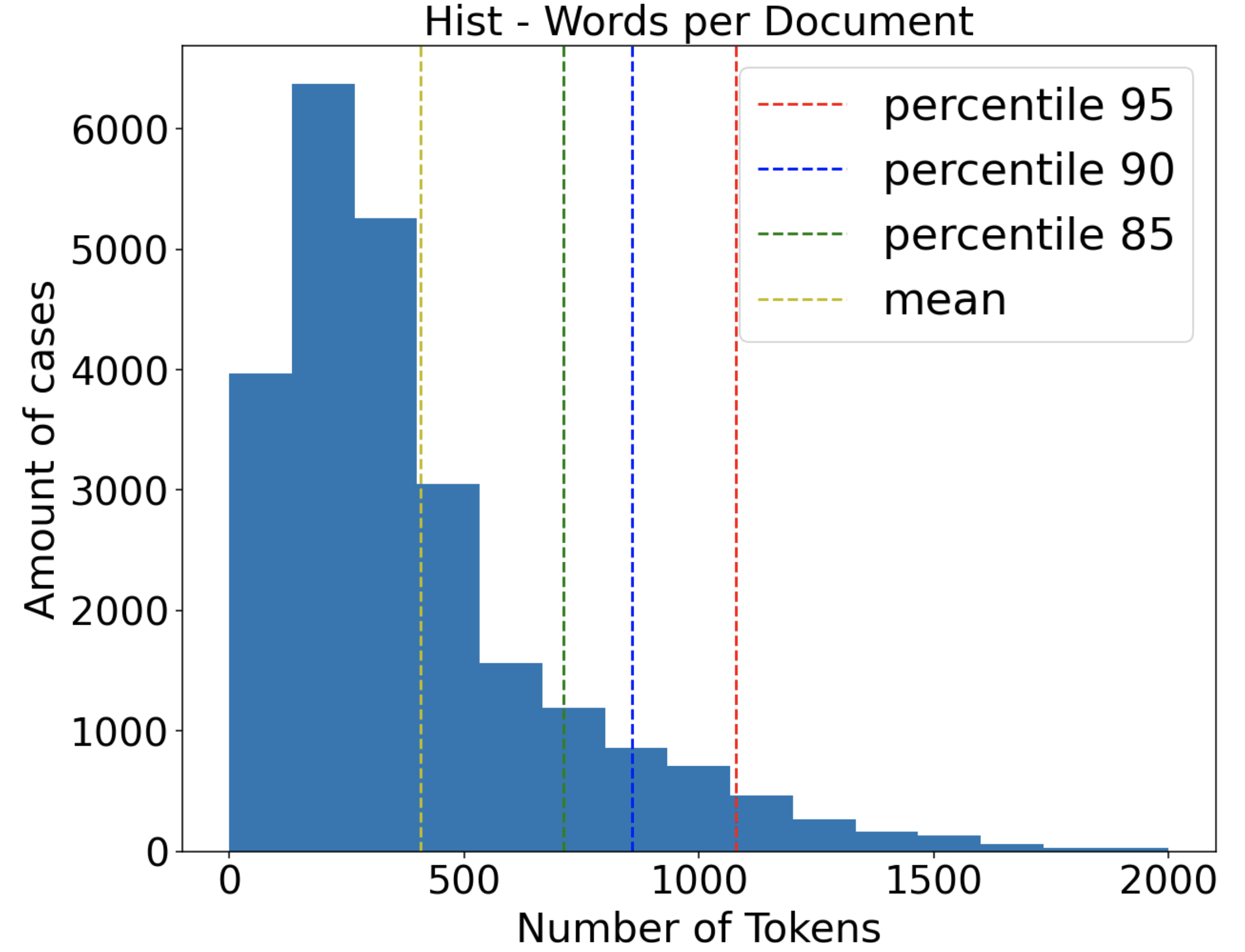} 
    \label{fig:separated_allegations_hist}
    }}
    }
    \caption{Histograms for the three dataset variants (number of tokens calculated using bert-base-uncased tokenizer).}
    \label{fig:histograms}
\vspace{-3mm}
\end{figure*}

We experimented with different variants of the dataset to study the effect of the different parts of the text. 
We deliberately focused our attention more on the allegations because the facts contain a lot of repetitive content and are noisier than the allegations (many paragraphs only contain citations). Additionally, the facts contain many citations to laws, which are less relevant to the case's outcome according to domain experts (the facts are more generic and less case-specific than the allegations).


\subsubsection*{Full Text}
The \emph{Full Text} dataset combines the plaintiff's facts and the allegations but also disregards any background details. We concatenated the facts at the beginning and added the allegations parts to create one input text.
We observe in Figure \ref{fig:full_text_hist} that this dataset is rather long -- almost 2700 tokens on average -- with 10\% of cases longer than 5400 tokens.

\subsubsection*{Unified Allegations}
The \emph{Unified Allegations} dataset consists of all case's allegations (mentioned in the complaint) concatenated together to form one input text . Approx. 2K documents did not contain any allegations (based on our extraction regexes), reducing the dataset size from 10.8K to 8.8K documents. 
The allegations make up a bit less than half of the full text complaint, as shown in Figure \ref{fig:unified_allegations_hist} (mean of $\sim$1,100 tokens and percentile 90 at $\sim$2,400 tokens).

\subsubsection*{Separated Allegations}
The \emph{Separated Allegations} dataset considers each allegation as a separate sample, increasing the size from 8.8K to 25K documents. We considered this dataset to test whether the entire context is necessary. 
Figure \ref{fig:separated_allegations_hist} shows the length distribution for individual allegations. Surprisingly, even a single allegation can reach up to 2,000 tokens ($\sim$ 4-5 pages of continuous text). However, most allegations (95\%) are not longer than roughly 2 pages (1,100 tokens) with the average at 400 tokens. 


\section{Experiments}

\subsection{Experimental Setup}
For all experiments, we truncated the text to the model's maximum sequence length (2,048 for Longformer and BigBird, 512 otherwise), unless otherwise specified. All experiments have been performed on the binarized labels (win or lose). 
We ran the experiments with 5-fold cross-validation and averaged across 5 random seeds. For more details regarding hyperparameter tuning and preprocessing, please refer to Appendix \ref{sec:additional_training_details}.

\subsection{Methods}

We compared the following pretrained transformer models: 
BERT
\cite{devlin_bert_2019}, 
LegalBERT
\cite{chalkidis_legal-bert_2020} (pretrained on diverse English legal data from Europe and the US with a domain-specific tokenizer), 
CaseLawBERT
\cite{zheng_when_2021} (pretrained on 37GB of US state and federal caselaw with a domain specific tokenizer),
LegalRoBERTa\footnote{\url{https://huggingface.co/saibo/legal-roberta-base}} (continued pretraining from RoBERTa checkpoint on 4.6 GB of US caselaw and patents), 
BigBird
\cite{zaheer_big_2021} 
and Longformer
\cite{beltagy_longformer_2020}. 
For all models, we used the publicly available base checkpoints on the Huggingface hub\footnote{\url{https://huggingface.co/models}}. We ran our experiments with the Huggingface transformers library \cite{wolf_transformers_2020} available under an Apache-2.0 license.
\subsection{Results}

\begin{table}[t]
\centering
\resizebox{0.9\columnwidth}{!}{
\begin{tabular}{lr}
\toprule
    Method          &  Accuracy  \\
\midrule
\multicolumn{2}{l}{Full Text (trunc. to 2048 tokens)}\\
\midrule
 Longformer         & $62.87 _{ \pm 2.06}$ \\
    BigBird         & $63.26 _{ \pm 3.40}$ \\
\midrule
\multicolumn{2}{l}{Unified Allegations (trunc. to 512 tokens)}\\
\midrule
            BERT    & $65.06 _{ \pm 1.67}$ \\
       LegalBERT    & $65.57 _{ \pm 0.26}$ \\
 CaseLawBERT        & $65.87 _{ \pm 0.60}$ \\
 LegalRoBERTa       & $65.95 _{ \pm 0.98}$ \\
\midrule
\multicolumn{2}{l}{Separated Allegations (trunc. to 512 tokens)}\\
\midrule
             BERT   & $64.98 _{ \pm 1.08}$ \\
        LegalBERT   & $65.57 _{ \pm 0.62}$ \\
  CaseLawBERT       & $66.82 _{ \pm 0.78}$ \\
 LegalRoBERTa       & $65.97 _{ \pm 0.88}$ \\
\bottomrule
\end{tabular}
}
\caption{Longformer and BigBird used a maximum sequence length of 2,048 tokens. All other models used 512 tokens. For all datasets, we truncated the text to fit the maximum sequence length.}
\label{tab:acc_all_truncation}
\vspace{-3mm}
\end{table}

Results are reported in the $mean _{ \pm std}$ format averaged accuracy across 5 random seeds. 
Table \ref{tab:acc_all_truncation} shows the main results. 
We observe that the setup considering the entire text is harder than when we only consider the allegations (best Full Text model is at $\sim$ 63\% and worst allegations model is at $\sim$ 65\%). These findings confirm our hypothesis, that the allegations encode more useful information than the facts (see Section \ref{sec:dataset_variants}) (the facts are often at the beginning of the complaints; thus the  models on the Full Text dataset are likely to see mostly facts because of the truncation).  



In line with previous findings \cite{chalkidis_lexglue_2021,chalkidis_legal-bert_2020, zheng_when_2021}, models with legal pretraining outperform BERT also in our datasets (Unified Allegations and Separated Allegations). However, for LegalBERT the difference is small (only 0.5\% above BERT). The models pretrained mostly or exclusively on US caselaw (LegalRoBERTa or CaseLawBERT respectively) perform better (up to 2\% better than BERT), presumably because our dataset also originates from the US. CaseLawBERT achieves a much higher difference to BERT on the CaseHOLD task (4.6 F1) \cite{zheng_when_2021} and on SCOTUS (7.6 macro-F1) \cite{chalkidis_lexglue_2021}. Both of these tasks are based on the same data as has been used in the pre-training of LegalRoBERTa and CaseLawBERT, whereas the complaints in our dataset are unseen by all models during pre-training. We suspect that this different data is the reason for the legal models not outperforming BERT as strongly as has been observed in other datasets.

\section{Error Analysis}
\acp{NN} and their latest incarnation, Transformers \cite{vaswani_attention_2017}, work very well across a wide range of tasks, especially if the tasks involve more ``complicated'' data like text or images. 
However, in contrast to traditional \ac{ML} methods such as Linear Regression, they are not interpretable out-of-the-box. 
Neural Networks need additional methods to make them explain themselves better to humans. 
A rich body of literature investigates how to make \acp{NN} and especially Transformers more interpretable \cite{ribeiro_why_2016, sundararajan_axiomatic_2017, lundberg_unified_2017, dhamdhere_how_2018, serrano_is_2019, bai_why_2021}.
Interpretability is especially important in high-stakes domains such as law or medicine.

In the following two sections, we analyze our models using the two interpretability methods Calibration and \ac{IG} to get a better understanding of their inner workings.

\subsection{Calibration}
\label{sec:calibration}


In this section, we investigate to what extent our models are calibrated out-of-the-box and ``calibratable''. Calibration is a first step towards understanding whether the model output can be trusted \cite{guo_calibration_2017, desai_calibration_2020}: how aligned are the confidence scores with the actual empirical likelihoods? Thus, if the model assigns 60\% probability to a label, then this label should be correct in 60\% of cases if the model is calibrated. So, even if the model itself is a black-box, a calibrated model at least gives an indication whether it knows when it is wrong. This information can be very valuable when deploying models in the real world because it allows us to discard predictions where the model is below some certainty threshold.
Well calibrated models are especially important in domains with high potential downside for users, such as predictive tools for court cases.

\begin{figure}[ht]
    \centering
    \subfloat[\centering Before Calibration]{{\includegraphics[width=\columnwidth]{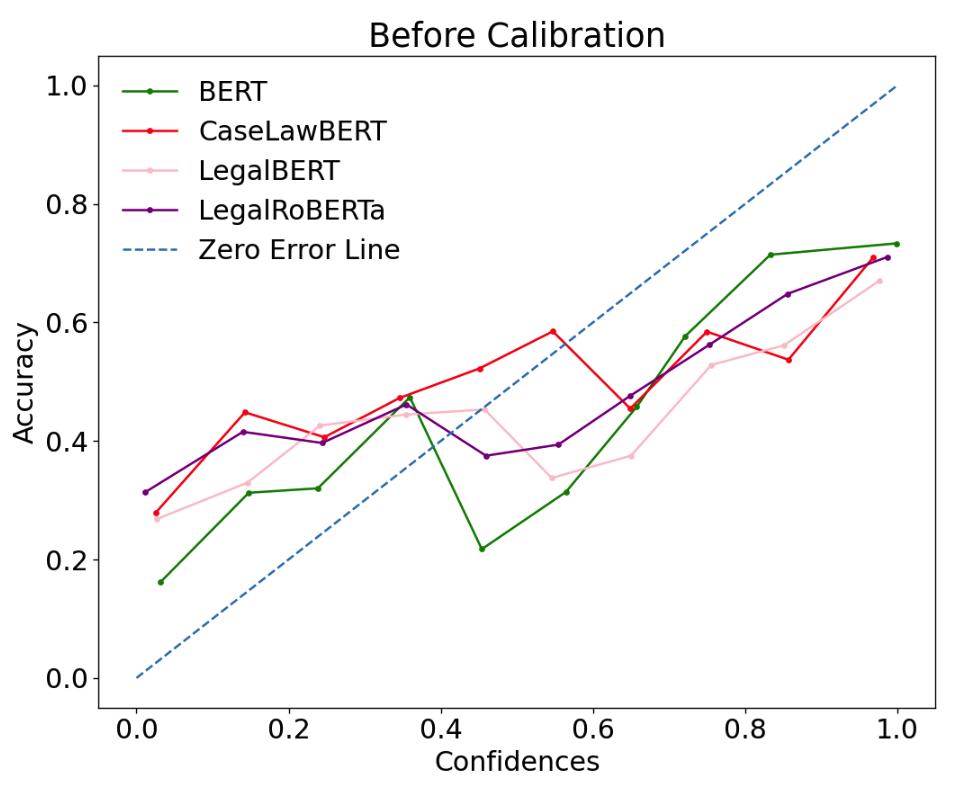} }}
    \qquad
    \subfloat[\centering After Calibration]{{\includegraphics[width=\columnwidth]{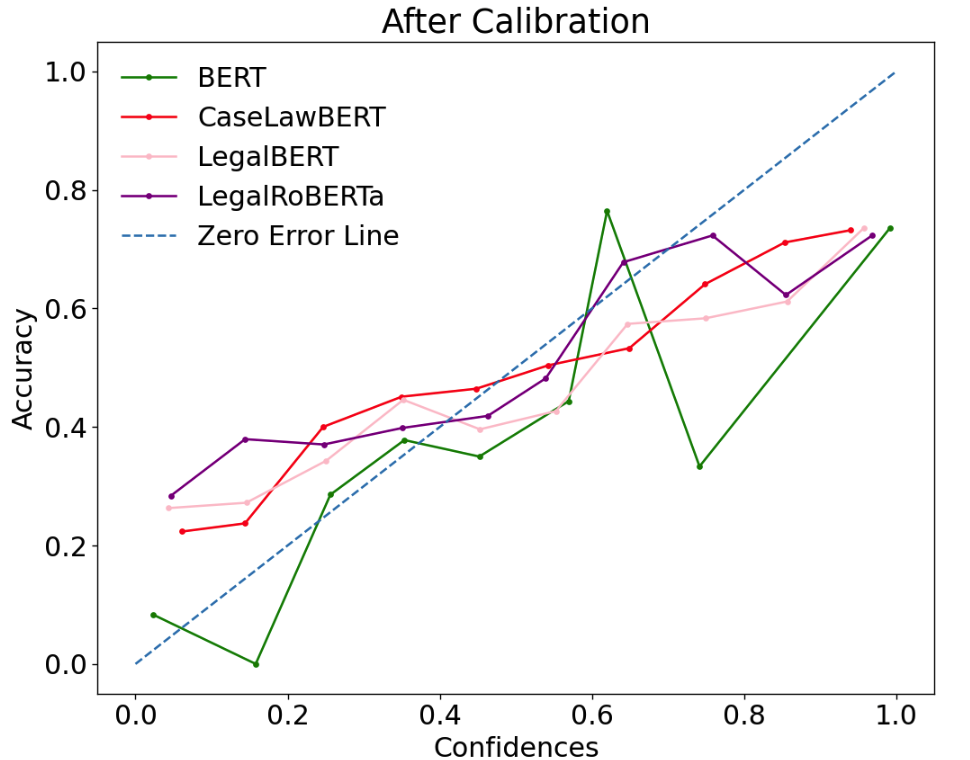} }}
    \caption{Calibration on the Unified Allegations dataset.}
    \label{fig:calibration}
    \vspace{-3mm}
\end{figure}

In this work, we follow \citet{desai_calibration_2020} by employing \ac{TS} \cite{guo_calibration_2017} for calibrating our models using the netcal library\footnote{\url{https://github.com/fabiankueppers/calibration-framework}} \cite{Kueppers_2020_CVPR_Workshops} available under an Apache License 2.0 license.
We show calibration plots in Figure \ref{fig:calibration} for BERT and the legal models on the Unified Allegations dataset and aggregated scores in Table \ref{tab:calibration} in Appendix \ref{sec:calibration_results}.
We observe that the legal models are less calibrated than BERT before, but better calibrated after \ac{TS}. So \ac{TS} seems to calibrate domain-specific models better than general models. When comparing the calibration of our models with the calibration of models from the literature \cite{desai_calibration_2020}, we note that our models are less calibrated overall (further away from the zero-error-line and higher ECE scores), both out-of-the-box and after applying \ac{TS}. We hypothesize that the generally lower accuracy on our hard dataset also makes the models less calibrated, especially in the areas of high (> 0.8) and low (< 0.2) confidence.  
To the best of our knowledge, in legal \ac{NLP} we are the first to perform such an analysis.

\subsection{Integrated Gradients}

\begin{figure*}[!ht]
    \includegraphics[width=\textwidth]{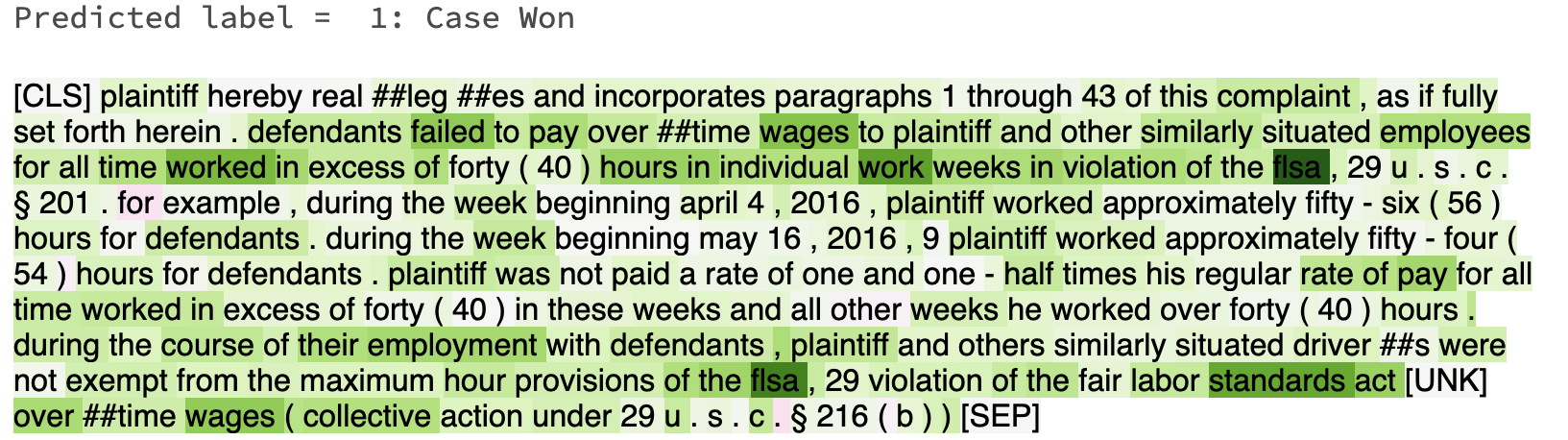}
    \caption{Analysis using \acf{IG}}
    \label{fig:integrated_gradients}
    \vspace{-3mm}
\end{figure*}

We conduct a qualitative analysis of the LegalBERT model using \ac{IG}\footnote{\url{https://github.com/cdpierse/transformers-interpret\#sequence-classification-explainer}} \cite{sundararajan_axiomatic_2017} and show an illustrative example in Figure \ref{fig:integrated_gradients}.
We observe that the model focuses most on ``flsa'' an acronym for Fair Labor Standards Act\footnote{\url{https://www.dol.gov/agencies/whd/flsa}} regulating minimum wage and overtime among others. Further, the model focuses on ``work'' and ``wages'' possibly signaling a (limited) understanding of the connections between those concepts. 
Future work may investigate explainability of \acp{PLM} in more detail on the \ac{LJP} task.

\section{Human Expert Annotations}

\citet{malik-etal-2021-ildc} collected predictions for the judgment outcome of Indian Supreme Court cases from five legal experts. The experts agreed with the judges in 94\% of the cases, on average. Note, however, that they have access to both the facts summary and the court's considerations. Their best model, XLNet + BiGRU, only achieves an accuracy of 78\%. 
Contrarily, \citet{jacob_de_menezes-neto_using_2022} find that all their models outperform 22 highly skilled experts on \ac{LJP} in Brazilian Federal Courts using the entire case description for prediction.

We asked legal experts (employees of our company) and US law students in their final year, to predict the judgment outcome of 200 randomly selected examples in our Full Text dataset. 
Note that they only had access to the facts and allegations from the plaintiff's pleas (same as our models), and not to the court case written by the judge. 
So, their task was much more difficult than the one posed to the annotators by \citet{malik-etal-2021-ildc} and \citet{jacob_de_menezes-neto_using_2022}. 
In our task, participants (whether models or human experts) basically need to estimate how the court is going to decide based only on the plaintiff's pleas. 
For each document, our legal experts had to answer whether they think the plaintiff would win or lose the case. 
Furthermore, they also had to indicate their confidence level for being correct (from 1 -- very unsure -- to 5 -- very sure). 
We made sure that the annotators did not look for any additional information regarding the complaint (e.g., news articles about the outcome or further information on different legal platforms) so that their answer is based only on the input text presented on the annotation platform. 
Figure \ref{fig:annotation_plattform} in Appendix \ref{sec:annotation} presents a screenshot of the annotation platform we used. 

On the entire dataset sample (200 examples), the human experts achieve an accuracy of 53\%. When we filtered out the samples where the human experts were not confident (confidence score 1, 2 or 3), they achieved an accuracy of 60\%. The entire results for the human experts are shown in Appendix \ref{sec:human_results} in Table \ref{tab:human_results}. 
We also trained and evaluated a Longformer model for comparison with the human predictions. We randomly split our remaining dataset into 6,877 train and 1,851 validation examples. 
Surprisingly, the Longformer model outperforms the human expert predictions both on the entire annotated test dataset (63\% vs. 53\% Accuracy) and the dataset filtered for high human confidence (67\% vs. 60\% Accuracy). In contrast to the human experts, the Longformer model only had access to the first 2,048 tokens of the case. While the human performance increases more than the Longformer performance on the high-confidence dataset, the Longformer model also has a higher performance, suggesting that these cases are easier to predict.

The task proposed in our dataset seems very challenging, given that human experts face great challenges in solving it. Interestingly, on the Indian dataset the humans clearly outperform the models, whereas in the Brazilian dataset it is reversed, similar to our results. Note that lawyers are often specialized in very narrow domains (legal areas). The cases in our dataset may be very diverse, and thus a generic model might be better suited for this task than specialized human experts. Future work may investigate this finding in more detail.


Figure \ref{fig:human_calibration} shows the calibration plot on the Full Text dataset, comparing Longformer before and after calibration with the human confidence scores. We observe that Longformer is already well calibrated in comparison to the human experts. Using \ac{TS}, the \ac{ECE} of Longformer can be reduced from 5.14 to 2.34, whereas the \ac{ECE} of the human experts lies at 17.5. Again, as mentioned in Section \ref{sec:calibration}, the lower accuracy of the humans might explain their worse calibration compared to Longformer.

\section{Conclusions and Future Work}

\subsection*{Answers to the Research Questions}

\noindent \textbf{RQ1}: \emph{To what extent is it possible to determine the winner of US class action cases using only the textual part of the complaints (without metadata)?}
It is possible, to some extent, to determine the winner of US class action cases using only the textual part of the complaints. Our best model achieves an accuracy of 66.8\% (LegalRoBERTa) on the datasets using only the allegations. However, as this number shows, there is still a lot of room for improvement.

\noindent \textbf{RQ2}: \emph{To what extent can we use \acf{TS} to better calibrate our models?}
Similar to Natural Language Inference, Paraphrase Detection and Commonsense Reasoning tasks \cite{desai_calibration_2020}, we also find that in the \ac{PJP} task, \ac{TS} helps in calibrating pretrained transformers. In our best model, \ac{TS} led to a decrease in \ac{ECE} scores from 28 to 2.

\noindent \textbf{RQ3}: \emph{To what extent can expert human lawyers solve the proposed task?}
Expert human lawyers perform better than chance on a randomly selected dataset of 200 samples and can increase their accuracy from 53\% to 60\% when they are confident in their decision. However, they are still outperformed by a Longformer model having access to only the first 2,048 tokens in both scenarios.


\subsection*{Conclusions}

We release a challenging new dataset of class action lawsuits for the more realistic \ac{PJP} task (where the input is based on the complaints instead of the further processed facts summary written by the judge) in the US, a jurisdiction with the common law system.
Additionally, we calibrated our models using \ac{TS} and found that despite the relatively low accuracy (66\% for the best model), relatively low \ac{ECE} scores around 2 can be achieved.
Finally, we find that our Longformer model is 10\% more accurate than the human experts on our dataset despite having only access to the first 2,048 tokens of the case.

\subsection*{Limitations}

Our best model achieves an accuracy of 66\%. This may suggest that either the task posed in this dataset is very hard, or we did not optimize our models enough. The results achieved by the human experts suggests that the former is the case. However, we believe much more work is needed here.

Although we did some first efforts to interpret our model's outputs using Calibration and \ac{IG}, the literature knows a host of other explainability methods \cite{molnar2022}. We leave a more thorough qualitative analysis involving domain experts and explainability methods for future work.

Our experiments were performed only on relatively short input spans (512 tokens for allegations, and 2048 for full text). Longformer or BigBird support input spans until 4096 tokens. Another possibility is the use of hierarchical models, as employed for example by \citet{niklaus_empirical_2022, dai_revisiting_2022} that can also easily scale to 4096 tokens given the right hardware. With 4096 tokens, we could fully encode all allegations and almost 80\% percent of the full texts. We leave these investigations to future work.

\subsection*{Future Work}

Since the legal models outperformed BERT only to a small margin, we suspect that further pretraining \cite{gururangan_dont_2020} on in-domain data might further enhance the performance. Additionally, in future work, we plan to study the domain-specific PJP and whether domain-specific models are better than generic model or human experts.

Large \acp{PLM} have proved to be very strong few shot learners in many tasks \cite{brown_language_2020, chowdhery_palm_2022}. The use of such models may bring performance boosts also in our studied task. We leave experimentation using different prompting strategies for future work \cite{arora_ask_2022, wei_chain_2022, suzgun_challenging_2022}. 

We discovered through our analysis using IG that some legal domains have a strong correlation to a particular label.
To produce complaints with a higher success likelihood in court, future studies may examine the linguistic structure of successful allegations.


\section*{Ethics Statement}

The goal of this research is to achieve a better understanding of \ac{LJP} to broaden the discussion and aid practitioners in developing better technology for both legal experts and non-specialists. We believe that this is a crucial application area, where research should be done \cite{tsarapatsanis-aletras-2021-ethical} to improve legal services and democratize legal data, making it more accessible to end-users, while also highlighting (informing the audience on) the various multi-aspect deficiencies seeking a responsible and ethical (fair) deployment of legal-oriented technology.

In this direction, we study how we can best build our dataset to maximize accuracy of our models on the task. Additionally, we study the inner workings of the models using \acl{IG} and make sure that our models are calibrated. A well calibrated model outputs confidence probabilities in line with actual likelihoods, thus giving the users the possibility of discarding low-confidence predictions or at least treating them with caution.

Lawyers often perform the \ac{LJP} task by giving their clients advice on how high the chances for success are in court for specific cases. Given the complaint documents, we were able to show in this work that our models outperformed human experts in this task.

But, like with any other application (like content moderation) or domain (e.g., medical), reckless usage (deployment) of such technology poses a real risk. According to our opinion, comparable technology should only be used to support human specialists (legal scholars, or legal professionals).


\section*{Acknowledgements}
We thank all the anonymous reviewers for their insightful comments. We thank the two employees at Darrow for the annotation of the dataset.

\bibliography{anthology,custom,references}
\bibliographystyle{acl_natbib}

\appendix

\section{Additional Training Details}
\label{sec:additional_training_details}

\subsection{Hyperparameter Tuning}
We randomly split the data into 70\% train, 15\% validation and 15\% test split. We searched the learning rate in \{1e-6, 5e-5, 1e-5\} and had the best results with 1e-5. We searched dropout in \{0, 0.001, 0.1, 0.2\} and finally chose 0. We searched the batch size in \{16, 32, 64\} and chose 16. Where GPU memory was not sufficient, we used gradient accumulation for a total batch size of 16. We searched the activation function in \{Relu, SoftMax, LeakyRelu\} and chose SoftMax. We searched weight decay in \{0, 0.1\} and found 0 to perform best. We used AMP mixed precision training and evaluation to reduce costs. We used early stopping on the validation loss with patience 2. If early stopping was not invoked, we trained for a maximum of 10 epochs.
We used an AWS EC2 G5 instance with 4 CPU cores, 16 GB RAM and one NVIDIA A10G GPU (24 GB of GPU memory)

\subsection{Preprocessing}
We experimented with the following preprocessing methods: (a) removing punctuation; (b) removing numerals; (c) stemming; (d) lemmatization; and (e) entity masking (e.g., ``Plaintiff James won would receive 30\% from the 3 million compensation fund'' $\rightarrow$  ``PERSON won would receive PERCENT from the MONEY compensation fund''). We found that only stemming improved the results.

\subsection{Training Times}
On the Unified Allegations dataset, training took approximately one hour for all the investigated models. On the Separated Allegations dataset, it took approximately two hours per model. On the Full Text dataset, it took approximately six hours for Longformer and approximately eight hours for BigBird. All training times are counted for five folds and one random seed on an AWS EC2 G5 instance with 4 CPU cores, 16 GB RAM and one NVIDIA A10G GPU (24GB of GPU memory).

\subsection{Library Versions}
We used the following libraries and associated versions:
python                            3.8,
transformers                      4.17.0,
xgboost                           1.5.2,
torch                             1.11.0+cu113,
tokenizers                        0.12.1,
spacy                             3.2.3,
scikit-learn                      1.1.1,
pandas                            1.3.4,
numpy                             1.20.3,
netcal                            1.2.1,
nltk                              3.6.5,
optuna                            2.10.1,
matplotlib                        3.4.3.

\section{Additional Results}
\label{sec:additional_results}

\subsection{Filtering the Datasets}

\begin{table}[t]
\centering
\resizebox{\columnwidth}{!}{
\begin{tabular}{lrr}
\toprule
    Method         & Max Seq Len &  Accuracy  \\
\midrule
Full Text \\
\midrule
 Longformer        & 2048      & $63.64 _{ \pm 0.72}$ \\
    BigBird        & 2048      & $62.00 _{ \pm 1.08}$ \\
\midrule
Separated Allegations \\
\midrule
             BERT  & 512       & $64.82 _{ \pm 1.73}$ \\
  CaseLawBERT      & 512       & $66.06 _{ \pm 0.84}$ \\
        LegalBERT  & 512       & $64.57 _{ \pm 1.89}$ \\
 LegalRoBERTa      & 512       & $65.41 _{ \pm 1.09}$ \\
\bottomrule
\end{tabular}
}
\caption{Longformer and BigBird used a maximum sequence length of 2,048 tokens. All other models used 512 tokens. For all datasets, we filtered out the rows larger than the maximum sequence length.}
\label{tab:acc_all_filter}
\end{table}

In Table \ref{tab:acc_all_filter} we show results for the Filter setup, where we filtered out texts containing more tokens than the maximum sequence lengths of the models used. We note that the results don't change significantly in comparison to Table \ref{tab:acc_all_truncation} (Truncation setup).

\subsection{XGBoost}
\label{sec:xgboost}

\begin{table}[t]
\centering
\resizebox{\columnwidth}{!}{
\begin{tabular}{lrl}
\toprule
              Method    &  Max Seq Len &             Accuracy \\
\midrule
Full Text \\
\midrule
       BERT              &            512 &  $60.40 _{ \pm 0.90}$ \\
 LegalBERT               &            512 & $61.79 _{ \pm 1.13}$ \\
        CaseLawBERT      &            512 &  $60.65 _{ \pm 0.32}$ \\
      LegalRoBERTa       &            512 & $60.37 _{ \pm 0.66}$ \\
    Longformer           &           2048 & $59.96 _{ \pm 1.24}$ \\
    BigBird              &           2048 & $60.98 _{ \pm 0.70}$ \\
\midrule
Unified Allegations \\
\midrule
       BERT              &            512 & $62.08 _{ \pm 0.71}$ \\
 LegalBERT               &            512 & $63.01 _{ \pm 0.60}$ \\
        CaseLawBERT      &            512 & $62.22 _{ \pm 0.59}$ \\
      LegalRoBERTa       &            512 & $62.32 _{ \pm 1.12}$ \\
    Longformer           &            512 &  $61.7 _{ \pm 0.82}$ \\
    BigBird              &            512 & $61.13 _{ \pm 1.02}$ \\
\midrule
Separated Allegations \\
\midrule
       BERT              &            512 &  $63.19 _{ \pm 0.49}$ \\
 LegalBERT               &            512 &  $64.17 _{ \pm 0.44}$ \\
        CaseLawBERT      &            512 & $63.81 _{ \pm 0.67}$ \\
      LegalRoBERTa       &            512 &  $64.52 _{ \pm 0.30}$ \\
    Longformer           &            512 &  $64.65 _{ \pm 0.40}$ \\
    BigBird              &            512 &  $63.38 _{ \pm 0.31}$ \\

\bottomrule
\end{tabular}
}
\caption{We fed the embeddings of the transformer models into an XGBoost \cite{chen_xgboost_2016}. For all datasets, we truncated the text to fit the maximum sequence length.}
\label{tab:xgboost}
\end{table}

Table \ref{tab:xgboost} shows the results for using XGBoost \cite{chen_xgboost_2016} on top of the embeddings instead of simple linear layers as it is reported in Table \ref{tab:acc_all_truncation}. We observe that this more sophisticated classification layer does not improve results.

\subsection{Calibration Results}
\label{sec:calibration_results}

\begin{table}[t]
\centering
\resizebox{\columnwidth}{!}{
\begin{tabular}{lrrrr}
\toprule
          Method &  Opt. Temp. &        ECE Before &           ECE After & Accuracy \\
\midrule
            BERT & $0.19 _{ \pm 0.03}$ & $23.44 _{ \pm 3.20}$ & $5.06 _{ \pm 1.96}$ & $65.06 _{ \pm 1.67}$ \\
 CaseLawBERT & $0.20 _{ \pm 0.03}$ & $25.67 _{ \pm 2.32}$ & $2.59 _{ \pm 0.90}$ & $65.57 _{ \pm 0.60}$ \\
       LegalBERT & $0.22 _{ \pm 0.02}$ & $24.78 _{ \pm 1.13}$ & $3.06 _{ \pm 1.78}$ & $65.87 _{ \pm 0.26}$ \\
LegalRobertaBase & $0.13 _{ \pm 0.02}$ & $28.02 _{ \pm 2.16}$ & $1.92 _{ \pm 0.85}$ & $65.95 _{ \pm 0.98}$ \\
\bottomrule
\end{tabular}
}
\caption{Calibration results on the Unified Allegations dataset. The text was always truncated to fit the model's maximum sequence length of 512 tokens. Opt. Temp. abbreviates the optimal temperature used for calibrating the models.}
\label{tab:calibration}
\end{table}

Table \ref{tab:calibration} shows the detailed aggregated \ac{ECE} scores together with the optimal temperature and the accuracy on the Unified Allegations dataset.

\subsection{Human Results}
\label{sec:human_results}

\begin{table}[t]
\centering
\resizebox{\columnwidth}{!}{
\begin{tabular}{lrrrr}
\toprule
             &  Precision &   Recall &  F1-score &  \# Examples \\
\midrule
All Results\\
\midrule
        lose &   49.41 & 45.65 &  47.45 &   92 \\
         win &   56.52 & 60.18 &  58.29 &  108 \\
    accuracy &   -     & -     &  53.50 &  200 \\
\midrule
High Confidence\\
\midrule
        lose &   75.00 & 37.50 &  50.00 &     24 \\
         win &   54.54 & 85.71 &  66.66 &     21 \\
    accuracy &   -     & -     &  60.00 &     45 \\
\bottomrule
\end{tabular}
}
\caption{Results of the human experts on the 200 randomly selected cases. Under High Confidence we show the results for only the examples where the human experts rated their confidence at 4 or 5 out of 5.}
\label{tab:human_results}
\end{table}

Table \ref{tab:human_results} shows the results of the human experts on the 200 randomly selected examples. 

\section{Annotation Platform}
\label{sec:annotation}

\begin{figure*}[!ht]
\includegraphics[width=\textwidth]{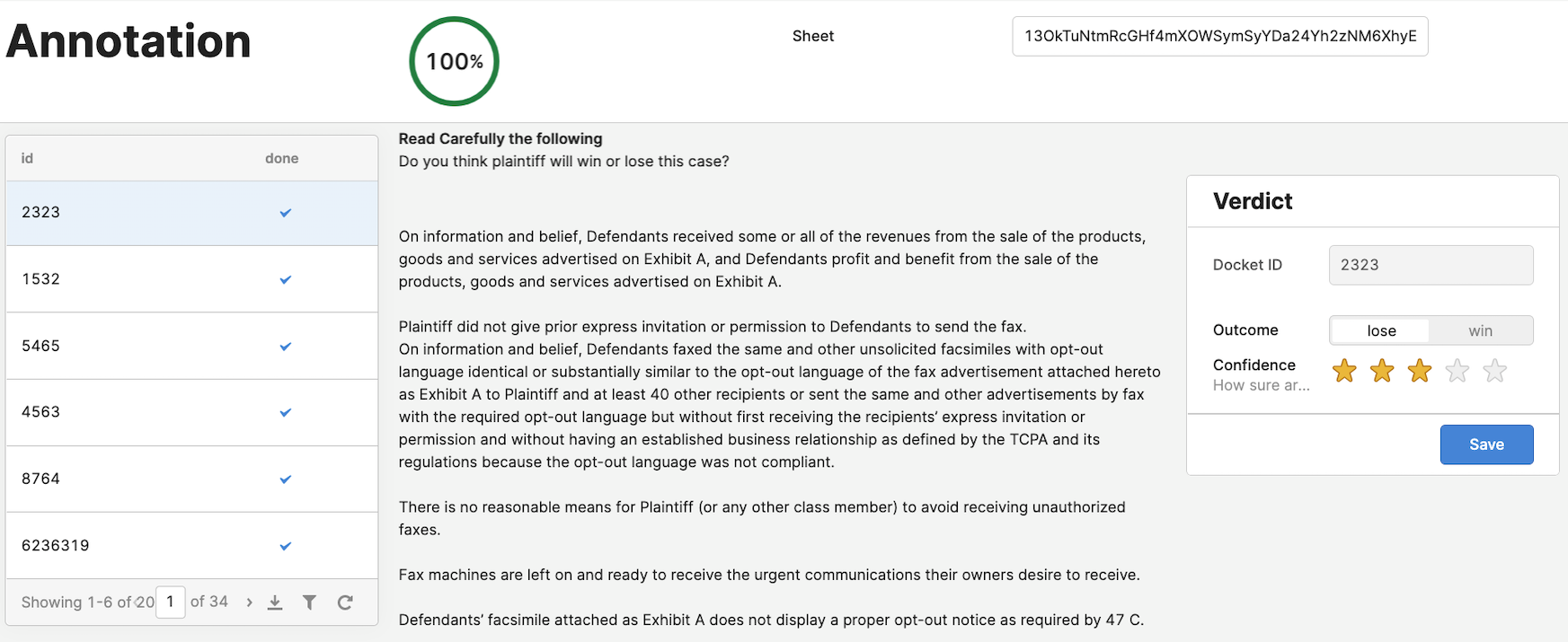}
\caption{The platform for the human annotations.}
\label{fig:annotation_plattform}
\end{figure*}

Figure \ref{fig:annotation_plattform} shows a screenshot of the annotation platform our human experts used.

\section{Example Complaint}
\label{sec:example_complaint}

\begin{figure}[!ht]
\includegraphics[width=\columnwidth]{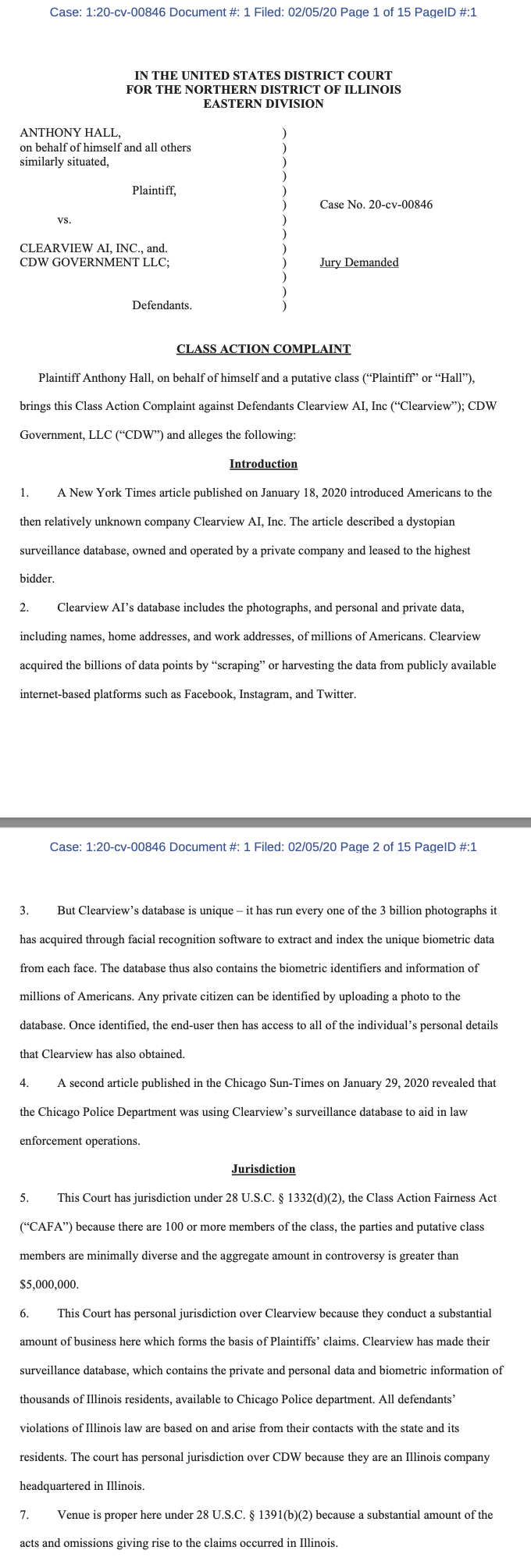}
\caption{These are the first two pages from an example complaint.}
\label{fig:example_complaint_first}
\end{figure}

\begin{figure}[!ht]
\includegraphics[width=\columnwidth]{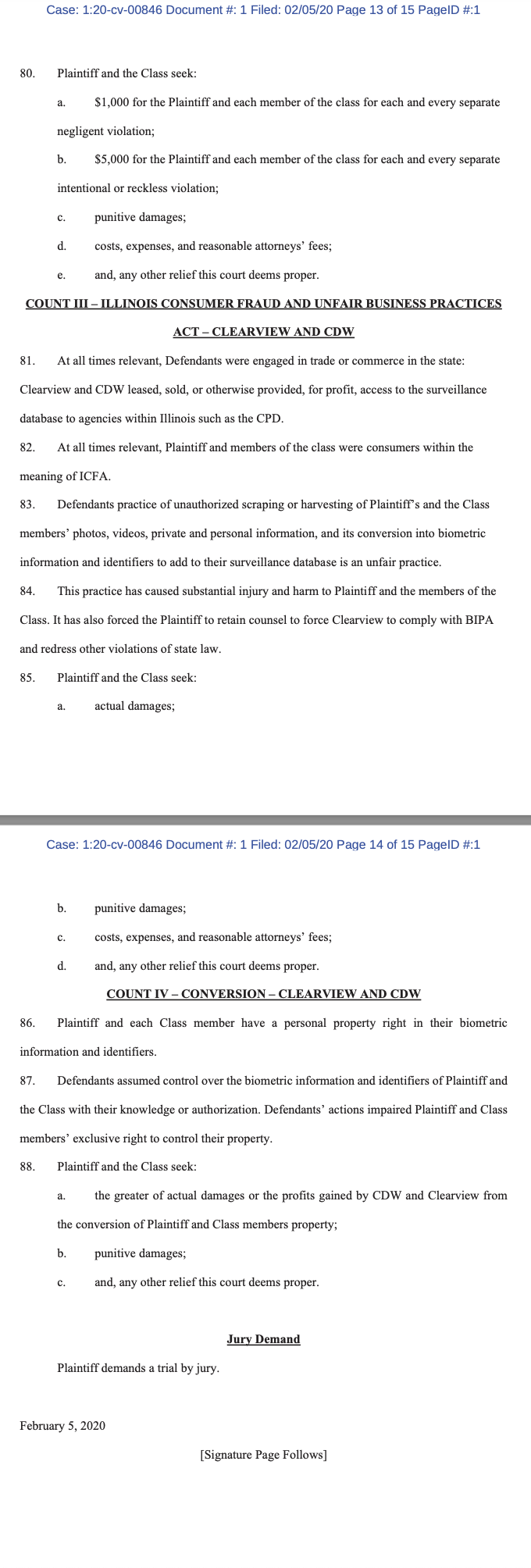}
\caption{These are the last two pages from an example complaint.}
\label{fig:example_complaint_last}
\end{figure}

Figures \ref{fig:example_complaint_first} and \ref{fig:example_complaint_last} show an example of a complaint present in the dataset.

\end{document}